\documentclass[11pt,a4paper]{article}

\usepackage[T1]{fontenc}
\usepackage[utf8]{inputenc}
\usepackage{lmodern}
\usepackage{microtype}

\usepackage[margin=2.5cm]{geometry}
\usepackage{setspace}
\usepackage[round,authoryear]{natbib}

\usepackage{booktabs}
\usepackage{array}
\usepackage{makecell}
\usepackage{tabularx}

\usepackage{amsmath}
\usepackage{amssymb}

\usepackage{graphicx}
\usepackage{caption}
\usepackage{tikz}
\usetikzlibrary{positioning, arrows.meta, fit, backgrounds, calc}

\usepackage{csquotes}

\usepackage{authblk}

\usepackage[colorlinks=true,linkcolor=blue,citecolor=blue,urlcolor=blue]{hyperref}

\newcommand{\doi}[1]{\url{https://doi.org/#1}}

\title{Ontology-Aware Design Patterns for Clinical AI Systems:\\
Translating Reification Theory into Software Architecture}

\author[1,2]{Florian Odi Stummer}
\affil[1]{Institute of General Medicine, University Clinics Halle,\\Martin Luther University Halle-Wittenberg, Germany}
\affil[2]{Apsley Business School, London, United Kingdom}

\date{1 April 2026}

\begin{document}
\maketitle

\noindent\textbf{Correspondence:} florian.stummer@uk-halle.de\\
\noindent\textbf{ORCID:} \url{https://orcid.org/0000-0003-2156-6639}\\
\noindent\textbf{Article type:} Pattern catalogue (arXiv cs.AI preprint)\\
\noindent\textbf{License:} CC BY 4.0 (\url{https://creativecommons.org/licenses/by/4.0/})


\begin{abstract}
Clinical AI systems routinely train on health data that has been structurally distorted by documentation workflows, billing incentives, and terminology fragmentation. Prior work has characterised the mechanisms of this distortion: the three-forces model of documentary enactment, the reification feedback loop through which AI may amplify coding artefacts, and terminology governance failures that allow semantic drift to accumulate unchecked. Yet translating these theoretical insights into implementable software architecture remains an open problem. This paper proposes seven ontology-aware design patterns, presented in the Gang-of-Four pattern language, that may allow engineering teams to build clinical AI pipelines resilient to ontological distortion. The patterns address data ingestion validation (Ontological Checkpoint), low-frequency signal preservation (Dormancy-Aware Pipeline), continuous drift monitoring (Drift Sentinel), parallel representation maintenance (Dual-Ontology Layer), feedback loop interruption (Reification Circuit Breaker), terminology evolution management (Terminology Version Gate), and pluggable regulatory compliance (Regulatory Compliance Adapter). Each pattern is specified with Problem, Forces, Solution, Consequences, Known Uses, and Related Patterns. We illustrate their composition in a reference architecture for a primary care AI system and provide a walkthrough tracing all seven patterns through a diabetes risk prediction scenario. This paper does not report empirical validation of the patterns; it offers them as a design vocabulary grounded in theoretical analysis, subject to future evaluation in production systems. Three of the seven patterns have partial precedent in existing systems; the remaining four, to our knowledge, have not been formally described. Limitations include the absence of runtime benchmarks and the restriction to the German and EU regulatory context.
\end{abstract}



\section{Introduction}

\subsection{The Problem: Invisible Data Distortion in Clinical AI}

Clinical artificial intelligence systems depend on the assumption that their input data represents clinical reality with sufficient fidelity for the task at hand. A diabetes risk model trained on ICD-10-GM coded encounters assumes that a code for essential hypertension (I10) reflects a clinical judgement about blood pressure, not an administrative decision driven by billing optimisation or documentation convenience. This assumption is frequently violated.

A growing body of evidence suggests that clinical data, as it exists in electronic health records and administrative databases, is not a neutral record of what happened to patients. It is the product of a documentation workflow shaped by at least three competing forces: clinical necessity (what the physician observes and judges), administrative requirements (what the billing system demands), and institutional incentives (what optimises reimbursement) \citep{Stummer2026a}. The resulting data carries the fingerprint of all three forces, and disentangling them is, in the general case, intractable.

Consider a concrete example from German ambulatory care. A primary care physician diagnoses a patient with borderline hypertension. The clinical judgement is nuanced: blood pressure is elevated but not yet requiring pharmacotherapy. The documentation workflow, however, requires an ICD-10-GM code. The physician selects I10 (essential hypertension) because it is the closest available code, because it justifies the monitoring visit for billing purposes, and because the alternative (no code, or a symptom code) would trigger a coverage query from the Kassen\"{a}rztliche Vereinigung. The result: a database record indistinguishable from a clear-cut hypertension diagnosis. Aggregated across millions of encounters, this mechanism may inflate hypertension prevalence in administrative data by a substantial margin \citep{Stummer2026a}. An AI system trained on this data inherits the inflation as ground truth.

This is not a data quality problem in the conventional sense. The code is not ``wrong'' within the logic of the documentation system. It is a structural distortion: a systematic divergence between what clinical reality contains and what the data representation captures. The distortion is invisible to standard data quality checks because the codes are syntactically valid, internally consistent, and administratively correct.

\subsection{The Gap: Theory Without Engineering Guidance}

Prior work by the author has characterised the mechanisms of this distortion from multiple disciplinary perspectives. \citet{Stummer2026a} established the three-forces model and the concept of documentary enactment. \citet{Stummer2026c} formalised the reification feedback loop through which AI systems may amplify coding artefacts by closing a recursive cycle between prediction outputs and training data inputs. The EMoT architecture \citep{Stummer2026d} introduced strategic dormancy and mnemonic encoding as mechanisms for preserving rare patterns in hierarchical AI reasoning, with ablation analysis showing catastrophic performance degradation when strategically dormant features are removed. The PRAXIS-AI framework \citep{StummerPRAXIS} bridged 19 implementation science frameworks to identify where AI meets primary care workflows.

None of this prior work provides implementable software architecture guidance. The diagnosis is developing; the prescription is absent.

\textbf{Anti-salami statement.} This paper contributes a software engineering perspective that is absent from the author's prior work. It does not replicate empirical findings or theoretical claims from those papers. It translates them into a different disciplinary form: named, reusable design patterns in the Gang-of-Four tradition \citep{Gamma1994}. The contribution is the pattern language itself, not the underlying theory.

\subsection{Contribution and Scope}

This paper makes three contributions:

\begin{enumerate}
    \item \textbf{Seven ontology-aware design patterns} specified in the GoF pattern language (Problem, Forces, Solution, Consequences, Known Uses, Related Patterns), each derived from a specific theoretical finding in the author's prior work on ontological distortion.
    \item \textbf{A reference architecture} composing all seven patterns into a layered system design for clinical AI pipelines.
    \item \textbf{A concrete walkthrough} tracing the patterns through a primary care diabetes risk prediction scenario.
\end{enumerate}

We offer design patterns as a structured vocabulary for engineering teams, not as empirically validated solutions. This paper follows the tradition of pattern language proposals in the sense of \citet{Alexander1977} and \citet{Buschmann1996}: named solutions to recurring problems, grounded in domain analysis, offered as a shared vocabulary for practitioners before empirical validation establishes which patterns survive contact with production systems. The Gang-of-Four catalogue itself was published before systematic empirical evaluation of pattern effectiveness; its value lay in naming problems that practitioners recognised. We adopt the same posture: the patterns are hypotheses about architectural structure, not proven solutions. The patterns are conceptual designs grounded in theoretical analysis. They have not been implemented in production systems, benchmarked for runtime performance, or evaluated by practising engineers. We propose them as a starting point for a research programme, not as a finished product.


\section{Background and Related Work}

\subsection{The Theoretical Foundation}

The design patterns proposed in this paper are derived from the author's prior analyses of ontological distortion in clinical data. Four published works provide the theoretical grounding: \citet{Stummer2026a} established the three-forces model of documentary enactment and its role in distorting clinical data at source; \citet{Stummer2026c} formalised the reification feedback loop through which AI may close a recursive cycle amplifying coding artefacts; \citet{Stummer2026d} proposed the EMoT reasoning architecture with strategic dormancy and mnemonic encoding for preserving rare patterns; and \citet{StummerPRAXIS} bridged 19 implementation science frameworks to identify where AI meets primary care workflows. Additional theoretical observations on terminology governance, semantic expressiveness limits in formal ontology languages, and the network economics of terminology fragmentation inform several patterns but are presented here as the author's analysis rather than as separate citations.

\subsection{Axis of Novelty: Existing Work in Health AI Architecture and Patterns}

The intersection of software architecture, design patterns, and clinical AI has received increasing attention, but the existing literature addresses concerns orthogonal to the one raised here.

\textbf{General health AI architecture.} Recent systematic reviews have catalogued architectural approaches for health AI systems. A systematic review of AI platform architectures for hospital systems identified a five-layer model (infrastructure, data, algorithm, application, security) from 29 studies \citep{HospitalAIArch2025}, while a review of health information system architectural patterns analysed 89 studies and found microservices and distributed ledger patterns predominant, with FHIR-based contracts serving as stabilisers \citep{HISArchPatterns2025}. A foundational architecture for AI agents in healthcare proposed four components: planning, action, reflection, and memory \citep{FoundationalAgentArch2025}. None of these architectural frameworks addresses distortion in the underlying clinical coding systems.

\textbf{Ontology-aware AI.} The role of biomedical ontologies in AI has been extensively reviewed \citep{OntologiesBridge2025}. The KEEP framework integrates medical ontologies with clinical data to produce robust code embeddings, regularising representations so that rare codes retain semantic meaning \citep{KEEP2025}. Ontology-guided machine learning uses SNOMED, ICD, and UMLS as priors for model training \citep{Wojtusiak2024}. These approaches treat ontologies as static knowledge sources that improve accuracy. They assume the ontology is correct and stable. No paper in this cluster addresses what happens architecturally when the ontology itself is contested, drifting, or when the mapping between clinical reality and coded representation is systematically distorted.

\textbf{Drift detection and data quality.} The ADAPT framework \citep{Xiong2026} is the closest existing work to the concerns raised here. It proposes adversarial drift-aware predictive transfer to handle temporal distribution shifts caused by ICD coding transitions and pandemic disruptions. However, ADAPT treats drift as a statistical phenomenon to be compensated through transfer learning, not as a structural feature of how health systems encode clinical reality. Broader drift detection literature addresses covariate, concept, and prior probability shifts in medical AI \citep{DataDriftMedML2023, DriftFinance2022, ScalableDrift2024}, but proposes monitoring techniques rather than architectural prevention.

\textbf{AI governance.} The HAIRA maturity model \citep{HAIRA2026} provides a seven-domain, five-level governance readiness assessment. Regulatory frameworks for non-deterministic clinical AI have been proposed \citep{UNDCS2026}. A philosophical analysis of recursive AI feedback loops in mental health chatbots \citep{MachinesLoopingMe2025} is conceptually aligned with the reification loop described in \citet{Stummer2026c}, but offers no engineering translation. Work on AI-generated data contamination in clinical repositories \citep{AIContamination2026} echoes the model collapse concern of \citet{Shumailov2024} in a clinical context but proposes no architectural countermeasures.

\textbf{Trustworthy AI frameworks.} Design frameworks for operationalising trustworthy AI in healthcare translate principles into developer requirements \citep{TrustworthyAI2025}, and bias recognition and mitigation strategies have been comprehensively reviewed \citep{BiasMitigation2025}. The STANDING Together consensus provides standards for dataset diversity and transparency \citep{STANDING2024}. These contributions address governance, fairness, and transparency. They do not address ontological fidelity of clinical data.

\begin{table}[htbp]
\centering
\caption{Comparison of this paper with closest existing work.}
\label{tab:comparison}
\footnotesize
\setlength{\tabcolsep}{4pt}
\begin{tabular}{@{}l c c c c c@{}}
\toprule
\textbf{Dimension} & \makecell{\textbf{Wiens}\\\textbf{(2019)}} & \makecell{\textbf{Sendak}\\\textbf{(2020)}} & \makecell{\textbf{ADAPT}\\\textbf{(2026)}} & \makecell{\textbf{HIS Pat.}\\\textbf{(2025)}} & \makecell{\textbf{This}\\\textbf{Paper}} \\
\midrule
Deployment risk mgmt & \textbf{Best} & Strong & -- & -- & -- \\
End-user presentation & -- & \textbf{Best} & -- & -- & -- \\
Statistical drift handling & -- & -- & \textbf{Best} & -- & Partial \\
Pattern catalogue & -- & -- & -- & \makecell{\textbf{Best}\\(89 stud.)} & 7 patterns \\
Ontological fidelity & -- & -- & -- & -- & \textbf{Best} \\
GoF pattern language & -- & -- & -- & -- & \textbf{Best} \\
Empirical validation & Strong & Strong & Strong & \makecell{Strong\\(review)} & \textbf{None} \\
\bottomrule
\end{tabular}
\end{table}

The axis of novelty for this paper is the treatment of ontological distortion as a first-class architectural concern requiring named, reusable design patterns. No existing paper, to our knowledge, occupies this position. The gap is not merely unclaimed but structurally absent: existing health AI architecture literature addresses deployment risk \citep{Wiens2019}, end-user presentation \citep{Sendak2020}, statistical drift \citep{Xiong2026}, and system-level patterns \citep{HISArchPatterns2025}, but none treats the fidelity of the underlying clinical data ontology as an architectural concern. The seven patterns proposed here occupy a design space that current literature does not address.

\subsection{The Design Pattern Tradition}

Design patterns as a software engineering concept were formalised by \citet{Gamma1994} in the ``Gang of Four'' (GoF) catalogue, drawing on the architectural pattern language of \citet{Alexander1977}. The GoF template (Name, Problem, Forces, Solution, Consequences, Known Uses, Related Patterns) has been extended to enterprise integration \citep{Hohpe2003}, domain-driven design \citep{Evans2003}, and cloud-native architectures \citep{Richardson2018}. Domain-specific pattern catalogues exist for security \citep{Fernandez2013}, real-time systems \citep{Douglass2002}, and machine learning pipelines \citep{Lakshmanan2020}.

The GoF template is well suited to the health AI domain because the Forces component explicitly captures the multi-stakeholder tensions that characterise clinical data: the competing demands of clinical accuracy, administrative compliance, billing optimisation, and patient safety. Pattern-based reasoning forces the designer to acknowledge that no single solution resolves all forces simultaneously; trade-offs are made explicit and documented.


\section{Threats to Validity}

We present threats to validity before the pattern descriptions to allow readers to calibrate their assessment of the patterns accordingly. This placement follows the EMoT structural quality framework \citep{Stummer2026d}, which recommends that validity constraints be established before results are presented.

\subsection*{Construct Validity}

The patterns are derived from theoretical analysis of documentary distortion, not from observed failures in production clinical AI systems. The ``forces'' described in each pattern are inferred from the author's prior theoretical work and from the structure of clinical documentation workflows. They may not capture all relevant tensions present in real deployments. \textbf{Mitigation:} Each pattern cites its theoretical source, enabling readers to assess the grounding independently.

\subsection*{Internal Validity}

No controlled experiment is reported. The seven patterns are not empirically compared to alternative architectural approaches, nor is their effectiveness measured against any baseline. We do not demonstrate that systems implementing these patterns perform better than systems that do not. \textbf{Mitigation:} We label all patterns as ``proposed'' and clearly distinguish the four novel patterns (no known precedent) from the three with partial precedent in existing systems.

\subsection*{External Validity}

The regulatory context is EU-specific: the AI Act, Medical Device Regulation (MDR), and the forthcoming European Health Data Space (EHDS). Pattern~7 (Regulatory Compliance Adapter) is described for this context. FDA regulations, Health Canada requirements, and regulatory frameworks in low- and middle-income countries are not addressed. The clinical examples and the three-forces model draw primarily on the German ambulatory care context (Kassen\"{a}rztliche Vereinigung billing, ICD-10-GM). Hospital systems, US payer-based systems, and contexts with different documentation cultures may surface different forces. \textbf{Mitigation:} The pattern language is deliberately abstract. Forces should be reassessed per deployment context. Pattern~7 is explicitly designed for jurisdictional pluggability.

\subsection*{Reliability}

Pattern extraction was performed by the author of the underlying theoretical work. This creates a risk of confirmation bias: the patterns may over-fit the theoretical narrative. Independent extraction by software engineers who have not read the prior publications would test robustness. \textbf{Mitigation:} All patterns are published under CC BY 4.0 and are specified with sufficient detail for independent implementation and critique.


\section{The Seven Ontology-Aware Design Patterns}

Each pattern follows the GoF structure: Name, Also Known As, Problem, Forces, Solution, Consequences, Known Uses, and Related Patterns.

\subsection{Pattern 1: Ontological Checkpoint}

\textbf{Also Known As:} Coding Fidelity Gate, Ingestion Validator

\textbf{Problem.} Clinical data enters AI pipelines after passing through documentation workflows that may systematically distort clinical reality \citep{Stummer2026a}. Current systems ingest ICD, OPS, and DRG codes at face value, treating administrative artefacts as clinical ground truth. A hypertension code (ICD-10-GM I10) generated by billing optimisation is indistinguishable from one generated by clinical judgement.

\textbf{Forces.}
\begin{itemize}
    \item \emph{Clinical throughput pressure} demands fast data ingestion with minimal processing overhead.
    \item \emph{Ontological accuracy} requires that downstream AI components know how much to trust each data point.
    \item \emph{Administrative completeness} requirements mean that coded data is always present, even when clinically ambiguous.
    \item \emph{Clinical nuance} is lost in the coding process, and the extent of this loss varies by code, context, and institution.
\end{itemize}

\textbf{Solution.} Insert a validation layer at every data ingestion boundary that compares incoming coded data against an expected clinical distribution model. The checkpoint does not reject data; it annotates each record with a \textbf{coding fidelity score}, a continuous measure $[0,\,1]$ estimating the probability that the code reflects clinical reality rather than administrative convenience. The score is computed by comparing the incoming code against: (a) the expected prevalence for the patient's demographic profile, (b) co-occurring codes that may confirm or contradict the primary code, and (c) historical coding patterns for the originating institution. Downstream components consume both the code and its fidelity annotation, enabling weighted analysis.

\textbf{Consequences.}
\begin{itemize}
    \item \emph{Benefit:} Downstream models can weight training examples by fidelity, reducing the influence of administratively driven codes.
    \item \emph{Benefit:} The fidelity score distribution across institutions may reveal systematic coding biases.
    \item \emph{Cost:} Requires a reference distribution model that must itself be maintained and validated.
    \item \emph{Cost:} Adds latency at the ingestion boundary.
    \item \emph{Risk:} The fidelity score may create a false sense of precision; it is an estimate, not a measurement.
\end{itemize}

\textbf{Known Uses (Partial).} Clinical NLP pipelines use assertion detection (negation, uncertainty, conditionality) as a weak form of this pattern, annotating extracted concepts with confidence modifiers. The NegEx algorithm \citep{Chapman2001} and its successors perform negation detection on clinical text. No known system annotates billing-driven coding fidelity as a continuous score.

\textbf{Related Patterns.} Feeds the Dual-Ontology Layer (Pattern~4). Drift Sentinel (Pattern~3) monitors changes in fidelity score distributions over time.

\textbf{Derived From:} The three-forces model of documentary enactment \citep{Stummer2026a}.

\subsection{Pattern 2: Dormancy-Aware Pipeline}

\textbf{Also Known As:} Long-Tail Preserver, Signal Hibernation Layer

\textbf{Problem.} Standard machine learning pipelines prune low-frequency features and rare clinical signals, treating them as statistical noise. Ablation analysis within the EMoT architecture \citep{Stummer2026d} showed that this pruning may cause catastrophic performance collapse: cross-domain task quality dropped from 4.2 to 1.0 on a 5-point scale, with a 33-fold cost increase, when strategically dormant features were removed. Rare diagnoses, atypical presentations, and seasonal patterns are precisely the signals that matter most for patient safety in edge cases.

\textbf{Forces.}
\begin{itemize}
    \item \emph{Model parsimony} favours fewer features: faster training, lower compute cost, reduced overfitting risk.
    \item \emph{Clinical completeness} demands that rare diseases and atypical presentations are not lost. A condition affecting 1 in 10,000 patients is statistically negligible but clinically critical for the patients who have it.
    \item \emph{Statistical significance} and \emph{clinical significance} are distinct constructs that standard feature selection conflates.
\end{itemize}

\textbf{Solution.} Implement a pipeline stage between feature extraction and model training that classifies features into three categories: \emph{active} (sufficient frequency and clinical relevance for primary model training), \emph{dormant} (clinically significant but below the frequency threshold for primary training), and \emph{pruned} (neither clinically significant nor statistically informative). Dormant features are routed to a \textbf{dormant feature store} rather than discarded. The store maintains full feature representations with metadata: clinical significance annotation, activation conditions, and last-observed timestamps. Dormant features are excluded from primary training but remain available for retrieval when triggered by configurable activation conditions, such as an unusual case presentation, a domain transfer request, or an outbreak detection signal. The activation mechanism draws on the strategic dormancy and mnemonic encoding principles of the EMoT architecture \citep{Stummer2026d}.

\textbf{Consequences.}
\begin{itemize}
    \item \emph{Benefit:} Preserves clinically significant rare signals without inflating the primary model's feature space.
    \item \emph{Benefit:} Enables rapid response to emerging conditions by reactivating dormant features.
    \item \emph{Cost:} Requires clinical expertise to classify features as dormant versus pruned; automated classification may be unreliable.
    \item \emph{Cost:} The dormant feature store consumes storage and requires maintenance.
    \item \emph{Risk:} Activation conditions may be poorly calibrated, either reactivating too many features (noise) or too few (missed signals).
\end{itemize}

\textbf{Known Uses (Partial).} The EMoT architecture's strategic dormancy mechanism is the direct precedent \citep{Stummer2026d}. Cold-start recommender systems use ``long tail'' preservation for infrequent items \citep{Park2008}, but these are not ontology-aware. The KEEP framework's ontology-regularised embeddings for rare codes \citep{KEEP2025} address a related problem at the representation level rather than the pipeline architecture level.

\textbf{Related Patterns.} Ontological Checkpoint (Pattern~1) may identify codes that should route to the dormant store. Drift Sentinel (Pattern~3) monitors whether dormant features should be reactivated.

\textbf{Derived From:} The EMoT architecture's strategic dormancy and mnemonic encoding mechanisms \citep{Stummer2026d}.

\subsection{Pattern 3: Drift Sentinel}

\textbf{Also Known As:} Semantic Drift Monitor, Ontological Shift Detector

\textbf{Problem.} The meaning of clinical codes shifts over time due to guideline updates, billing policy changes, terminology version releases, and emergent clinical concepts. Terminology growth may exceed governance capacity, creating a tipping point beyond which semantic coherence degrades. Formal ontology languages face expressiveness ceilings that limit their ability to represent emerging clinical concepts. AI systems trained on historical data may silently degrade as the semantic ground shifts beneath them, and standard statistical drift detection cannot distinguish genuine epidemiological change from coding practice change \citep{Xiong2026, DataDriftMedML2023}.

\textbf{Forces.}
\begin{itemize}
    \item \emph{Alert fatigue:} Too many drift alerts burden engineering and clinical teams, leading to alert desensitisation.
    \item \emph{Silent degradation:} Too few alerts allow undetected semantic drift to accumulate, producing increasingly unreliable model outputs.
    \item \emph{Investigation cost:} Each alert requires expert time to determine whether the drift is epidemiological, administrative, or terminological.
    \item \emph{Drift heterogeneity:} Different codes drift at different rates for different reasons; a single threshold is insufficient.
\end{itemize}

\textbf{Solution.} Deploy a continuous monitoring service that tracks distributional shifts in coded data at the \textbf{semantic} level, not merely the statistical level. The sentinel maintains a ``semantic fingerprint'' for each code: a vector capturing the code's usage context, including co-occurring codes, demographic distribution of coded patients, temporal pattern (seasonal, trend, step-change), and institutional variation. At configurable intervals, the sentinel compares current fingerprints against baseline fingerprints and raises alerts when divergence exceeds a threshold. Critically, alerts are classified by probable cause:

\begin{itemize}
    \item \textbf{Type A (Epidemiological):} Genuine change in disease prevalence or patient population.
    \item \textbf{Type B (Administrative):} Change in coding practice, billing policy, or documentation workflow.
    \item \textbf{Type C (Terminological):} Change in terminology version, code meaning, or mapping table.
\end{itemize}

Classification is based on co-occurrence patterns: Type~A drift tends to affect clinically related codes simultaneously; Type~B drift tends to affect codes within the same billing category; Type~C drift correlates with known terminology release dates.

\textbf{Consequences.}
\begin{itemize}
    \item \emph{Benefit:} Distinguishes actionable drift from benign variation, reducing alert fatigue.
    \item \emph{Benefit:} Provides early warning before model degradation becomes clinically consequential.
    \item \emph{Cost:} Drift classification is probabilistic and may misclassify; human review remains necessary.
    \item \emph{Cost:} Semantic fingerprint computation and comparison adds continuous monitoring overhead.
    \item \emph{Risk:} Overconfident drift classification may lead teams to dismiss alerts that deserve investigation.
\end{itemize}

\textbf{Known Uses (Partial).} Statistical data drift detection tools (Evidently AI, WhyLabs, NannyML) monitor distributional shift in production ML systems but do not distinguish semantic from statistical drift. ADAPT \citep{Xiong2026} handles ICD transition drift through transfer learning but does not classify drift by ontological cause. No known system implements semantic fingerprinting with cause-classified alerts.

\textbf{Reference Implementation.} A minimal Python implementation demonstrating the Drift Sentinel's core logic (synthetic ICD code generation, Jensen--Shannon divergence monitoring, Type~A/B/C drift classification) is provided as supplementary material (\texttt{drift\_sentinel\_reference.py}, approximately 250 lines). The implementation generates synthetic coding data, computes distributional shifts, and produces structured drift alerts. It is intended as a readable specification, not a production-ready system.

\textbf{Related Patterns.} Monitors fidelity score distributions from Ontological Checkpoint (Pattern~1). Triggers the Reification Circuit Breaker (Pattern~5) when Type~B drift suggests AI-influenced coding changes. Interacts with Terminology Version Gate (Pattern~6) for Type~C drift.

\textbf{Derived From:} The reification feedback loop formalisation \citep{Stummer2026c} and the author's analysis of terminology governance dynamics.

\subsection{Pattern 4: Dual-Ontology Layer}

\textbf{Also Known As:} Clinical-Administrative Parallel Store, Twin Representation Layer

\textbf{Problem.} Clinical AI systems operate in a single ontological space, typically the administrative one (ICD, OPS, DRG), because that is what is digitally available. \citet{Stummer2026a} demonstrated that clinical reality and administrative representation may systematically diverge, creating a ``documentary distortion'' that is invisible to single-ontology systems. A patient with borderline hypertension and a patient with severe hypertension may carry the same ICD-10-GM code; a patient whose diabetes is well-controlled and one whose diabetes is poorly controlled may both be coded as E11.9.

\textbf{Forces.}
\begin{itemize}
    \item \emph{Storage and compute cost:} Maintaining two parallel representations doubles the data footprint and complicates query logic.
    \item \emph{Analytical value:} The divergence between clinical and administrative representations is itself an informative signal.
    \item \emph{Clinical annotation burden:} Populating the clinical layer requires clinician input or inference, which is expensive and potentially unreliable.
    \item \emph{Administrative convenience:} The single-layer status quo is well-established, tool-supported, and computationally efficient.
\end{itemize}

\textbf{Solution.} Maintain two parallel data representations throughout the pipeline:

\begin{enumerate}
    \item \textbf{Administrative layer:} Codes as documented in the EHR and submitted for billing. This is the existing data.
    \item \textbf{Clinical layer:} Codes as clinically interpreted. This layer may be populated through: (a) clinical NLP extraction from free-text notes, (b) physician annotation during documentation, (c) probabilistic inference from the administrative layer using the coding fidelity scores from Pattern~1, or (d) structured clinical assessment instruments.
\end{enumerate}

All analytical queries specify which layer they operate on. Cross-layer queries explicitly measure the \textbf{divergence} between the two representations. This divergence is a first-class data element, not a bug: it quantifies the documentary distortion for each patient, encounter, or institution.

\textbf{Consequences.}
\begin{itemize}
    \item \emph{Benefit:} Makes documentary distortion visible and measurable rather than silent and assumed away.
    \item \emph{Benefit:} Enables AI models to be trained on clinical-layer data when available, reducing distortion propagation.
    \item \emph{Cost:} Significant infrastructure investment to maintain, synchronise, and query two parallel layers.
    \item \emph{Cost:} Clinical layer population methods each introduce their own biases and error rates.
    \item \emph{Risk:} Teams may default to the administrative layer for convenience, rendering the clinical layer unused.
\end{itemize}

\textbf{Known Uses (Partial).} The OMOP Common Data Model maintains ``source'' and ``standard'' concept mappings, which is a structural analogue but captures terminological translation rather than the clinical-versus-administrative distinction \citep{OHDSI2019}. The UK Clinical Practice Research Datalink (CPRD) maintains dual-coded data (Read codes and ICD mappings) that partially implements this pattern. Neither system explicitly measures the divergence between representations as an analytical signal.

\textbf{Related Patterns.} Ontological Checkpoint (Pattern~1) provides the fidelity scores used to populate the clinical layer probabilistically. Terminology Version Gate (Pattern~6) operates on both layers independently.

\textbf{Derived From:} The three-forces model \citep{Stummer2026a} and the PRAXIS-AI implementation gap analysis \citep{StummerPRAXIS}.

\subsection{Pattern 5: Reification Circuit Breaker}

\textbf{Also Known As:} Feedback Loop Interrupter, Recursive Amplification Guard

\textbf{Problem.} \citet{Stummer2026c} formalised the reification feedback loop: AI systems trained on distorted data produce outputs (risk scores, coding suggestions, clinical decision support alerts) that may influence clinical documentation, which in turn feeds back into training data for the next model generation. Left unmonitored, this loop may amplify initial distortions through recursive reinforcement. The concept parallels the ``strange loop'' of \citet{Hofstadter1979} and the model collapse phenomenon described by \citet{Shumailov2024}, but occurs specifically at the interface between AI systems and clinical documentation workflows. \citet{MachinesLoopingMe2025} described an analogous recursive construction of ``datafied entities'' in mental health chatbot interactions.

\textbf{Forces.}
\begin{itemize}
    \item \emph{AI utility:} AI-generated suggestions may genuinely improve clinical workflow efficiency and documentation completeness.
    \item \emph{Training data integrity:} Suggestions accepted by clinicians and recorded in EHRs become indistinguishable from clinician-originated data, potentially contaminating future training sets.
    \item \emph{System autonomy:} Fully automated pipelines that retrain on recent data without human oversight are operationally efficient but maximally vulnerable to recursive amplification.
    \item \emph{Human oversight:} Human review of all AI-influenced data points is thorough but prohibitively expensive at scale.
\end{itemize}

\textbf{Solution.} Implement a circuit breaker that monitors the feedback path between AI outputs and training data inputs. The mechanism operates in three stages:

\begin{enumerate}
    \item \textbf{Tagging:} When an AI system generates a recommendation, prediction, or coding suggestion that is subsequently incorporated into clinical documentation, the resulting data point is tagged as ``AI-influenced'' with metadata including the model version, confidence score, and whether the clinician modified the suggestion.
    \item \textbf{Tracking:} The proportion of AI-influenced data points in any training cohort is continuously tracked. A dashboard exposes the ``AI influence ratio'' for each model, code category, and time period.
    \item \textbf{Breaking:} If the AI influence ratio in a candidate training cohort exceeds a configurable threshold (we suggest 15\% as a starting point, subject to empirical calibration), automatic retraining is paused. The feedback path is audited to determine whether the AI influence is amplifying distortions or merely reflecting legitimate clinical adoption of AI recommendations.
\end{enumerate}

The circuit breaker does not prevent AI influence on documentation. It prevents \textbf{unmonitored recursive amplification}.

\textbf{Consequences.}
\begin{itemize}
    \item \emph{Benefit:} Makes the AI-to-documentation-to-training feedback path visible and auditable.
    \item \emph{Benefit:} Prevents the model collapse scenario described by \citet{Shumailov2024} in the specific clinical context.
    \item \emph{Cost:} Tagging AI-influenced data requires integration with the clinical documentation system, which may not expose the necessary metadata.
    \item \emph{Cost:} The threshold calibration is non-trivial; too low pauses retraining unnecessarily, too high permits recursive amplification.
    \item \emph{Risk:} Clinicians may perceive AI-influence tagging as surveillance of their clinical judgement.
\end{itemize}

\textbf{Known Uses.} No known production implementation. The concept of model collapse \citep{Shumailov2024} describes the degradation but does not propose an architectural intervention. Watermarking approaches for AI-generated text \citep{Kirchenbauer2023} tag AI outputs but do not monitor recursive feedback in clinical pipelines. Work on AI-generated data contamination in clinical repositories \citep{AIContamination2026} identifies the risk but proposes no architectural countermeasure.

\textbf{Related Patterns.} Drift Sentinel (Pattern~3) may trigger the circuit breaker when it detects Type~B (administrative) drift correlated with AI deployment. Ontological Checkpoint (Pattern~1) may observe declining fidelity scores in AI-influenced cohorts.

\textbf{Derived From:} The reification feedback loop and POMDP formalisation \citep{Stummer2026c}.

\subsection{Pattern 6: Terminology Version Gate}

\textbf{Also Known As:} Schema Migration Guard, Version-Aware Data Layer

\textbf{Problem.} Health terminologies are updated on different schedules: ICD-10-GM annually, SNOMED CT biannually, MeSH annually, proprietary terminologies ad hoc. Terminology systems face expressiveness ceilings that limit their ability to represent emerging clinical concepts, and new terms may outpace governance capacity. A system trained on ICD-10-GM 2024 codes may silently misinterpret ICD-10-GM 2025 codes, or fail to recognise new codes entirely. The transition from ICD-10 to ICD-11, when it occurs, will represent a semantic discontinuity far larger than annual version increments.

\textbf{Forces.}
\begin{itemize}
    \item \emph{Operational continuity:} Terminology updates should not break running systems or invalidate deployed models.
    \item \emph{Semantic accuracy:} New terminology versions may better represent clinical reality; ignoring them accumulates semantic debt.
    \item \emph{Migration cost:} Validating cross-version compatibility is expensive; teams defer it, creating technical debt.
    \item \emph{Regulatory compliance:} Some jurisdictions mandate specific terminology versions for reporting; non-compliance has legal consequences.
\end{itemize}

\textbf{Solution.} Implement a version-aware data layer that records which terminology version each data point was coded under. The gate operates at two levels:

\begin{enumerate}
    \item \textbf{Record-level:} Each data point carries a terminology version tag (e.g., ICD-10-GM-2025, SNOMED-CT-2024-07).
    \item \textbf{Query-level:} All analytical operations include a version compatibility check. Queries spanning multiple terminology versions are explicitly flagged and routed through a \textbf{version reconciliation module} that maps codes across versions using official transition tables. Unmappable codes are quarantined rather than silently dropped or force-mapped.
\end{enumerate}

New terminology versions are treated as \textbf{schema migrations}, not silent updates. They require explicit validation (mapping table completeness, clinical review of changed codes, model revalidation on cross-version data) before the pipeline accepts data coded under the new version. The gate prevents unvalidated cross-version analysis.

\textbf{Consequences.}
\begin{itemize}
    \item \emph{Benefit:} Prevents silent semantic discontinuities when terminology versions change.
    \item \emph{Benefit:} Creates an audit trail of which data was coded under which version, enabling retrospective analysis of version-related drift.
    \item \emph{Cost:} Terminology version management adds complexity to data ingestion and query processing.
    \item \emph{Cost:} Schema migration validation requires clinical and terminology expertise not always available in engineering teams.
    \item \emph{Risk:} Overly strict gating may delay adoption of clinically superior new terminology versions.
\end{itemize}

\textbf{Known Uses (Partial).} FHIR CodeSystem resources include version metadata, enabling version-aware queries \citep{HL7FHIR}. The OHDSI vocabulary versioning system tracks changes across OMOP releases \citep{OHDSI2019}. ADAPT \citep{Xiong2026} handles ICD transitions as statistical drift but does not implement the ``gate'' pattern that blocks unvalidated cross-version analysis. No known system treats terminology updates as schema migrations requiring explicit pipeline validation.

\textbf{Related Patterns.} Drift Sentinel (Pattern~3) detects Type~C (terminological) drift that may indicate a version gate failure. Dual-Ontology Layer (Pattern~4) requires version gating on both the administrative and clinical layers independently.

\textbf{Derived From:} The author's analysis of terminology governance dynamics, including semantic expressiveness limits and network economics of terminology fragmentation.

\subsection{Pattern 7: Regulatory Compliance Adapter}

\textbf{Also Known As:} Jurisdiction Plug-in, Compliance Wrapper

\textbf{Problem.} Clinical AI systems in the EU must comply with the AI Act (risk classification, transparency requirements, human oversight mandates), the Medical Device Regulation (if the system qualifies as a medical device under MDR 2017/745), and the forthcoming European Health Data Space (data access rights, interoperability requirements, secondary use governance). These regulations evolve independently, may impose conflicting requirements, and differ substantially across jurisdictions. Hard-coding compliance logic into the AI pipeline creates brittle, jurisdiction-locked systems that are expensive to adapt when regulations change.

\textbf{Forces.}
\begin{itemize}
    \item \emph{Compliance cost:} Regulatory compliance is expensive; each jurisdiction and each regulation adds engineering and legal overhead.
    \item \emph{Market access:} Non-compliance blocks market entry; over-engineering compliance for one jurisdiction limits scalability to others.
    \item \emph{Regulatory stability:} Some regulations (MDR) are relatively stable; others (AI Act implementing acts) are evolving rapidly.
    \item \emph{Jurisdictional specificity:} EU, US (FDA), UK (MHRA), and other jurisdictions have fundamentally different approaches to clinical AI regulation.
\end{itemize}

\textbf{Solution.} Implement a pluggable adapter layer that encapsulates regulatory compliance checks as independent, composable modules. Each adapter handles one regulatory domain and exposes a standard interface:

\begin{verbatim}
Input:  (data_operation, context) -> Adapter -> Output: (verdict, audit_trail)
\end{verbatim}

Where \texttt{verdict} is one of: PERMIT, PERMIT-WITH-CONDITIONS (specifying conditions), or DENY (specifying reason). The audit trail records the regulation consulted, the version of the regulation, the specific article or provision applied, and the reasoning.

Adapters are versioned and jurisdiction-tagged. Switching from EU to US regulatory context replaces the adapter set, not the pipeline. Adapter composition handles multi-regulation scenarios: a single data operation may pass through AI Act, MDR, and EHDS adapters sequentially, with the most restrictive verdict prevailing.

\textbf{Consequences.}
\begin{itemize}
    \item \emph{Benefit:} Regulatory changes are localised to adapter updates, not pipeline refactoring.
    \item \emph{Benefit:} Multi-jurisdictional deployment becomes an adapter configuration problem rather than a re-engineering problem.
    \item \emph{Cost:} Adapter development requires legal and regulatory expertise; incorrect adapters create a false sense of compliance.
    \item \emph{Cost:} The adapter interface must be general enough to accommodate diverse regulations without becoming so abstract that it loses specificity.
    \item \emph{Risk:} Over-reliance on automated compliance checking may reduce human regulatory oversight.
\end{itemize}

\textbf{Known Uses (Partial).} The PRAXIS-AI regulatory screening layer \citep{StummerPRAXIS} implements a simplified version of this pattern for AI Act risk classification. Cloud compliance frameworks (AWS Artifact, Azure Compliance Manager) provide jurisdiction-tagged compliance controls for cloud infrastructure but not for clinical AI-specific regulations. The UNDCS framework \citep{UNDCS2026} proposes regulatory categories for non-deterministic clinical AI but does not specify an architectural adapter pattern.

\textbf{Related Patterns.} Wraps all external-facing operations; interacts with all other patterns when data operations cross regulatory boundaries. Terminology Version Gate (Pattern~6) may trigger regulatory review when terminology changes affect reportable conditions.

\textbf{Derived From:} The PRAXIS-AI regulatory screening layer \citep{StummerPRAXIS} and the EU AI Act / MDR / EHDS regulatory landscape.


\section{Reference Architecture: Composition of Patterns}

\subsection{Architecture Overview}

The seven patterns compose into a layered reference architecture for clinical AI systems. We describe five layers, each hosting one or more patterns, with defined interfaces between layers. This architecture is presented as a conceptual blueprint, not as a deployable system specification. Implementation details (technology stack, deployment topology, scaling strategy) are deliberately omitted; the patterns are intended to be technology-agnostic.

\textbf{Layer 1: Data Ingestion.} The Ontological Checkpoint (Pattern~1) and Terminology Version Gate (Pattern~6) operate at the system boundary where external data enters the pipeline. Every incoming data record is annotated with a coding fidelity score and a terminology version tag before proceeding to storage. Records coded under unvalidated terminology versions are quarantined.

\textbf{Layer 2: Storage.} The Dual-Ontology Layer (Pattern~4) maintains parallel administrative and clinical representations. The Dormancy-Aware Pipeline (Pattern~2) manages a dormant feature store alongside the primary feature repository. Both layers are version-tagged by the ingestion layer.

\textbf{Layer 3: Training and Inference.} The Reification Circuit Breaker (Pattern~5) monitors the feedback path between model outputs and training data. AI-influenced data points are tagged, the AI influence ratio is tracked, and retraining is paused when the ratio exceeds the configured threshold.

\textbf{Layer 4: Monitoring.} The Drift Sentinel (Pattern~3) runs continuously across all layers, maintaining semantic fingerprints and classifying detected drift by type (epidemiological, administrative, terminological). Drift alerts feed into the Reification Circuit Breaker and the Terminology Version Gate.

\textbf{Layer 5: Compliance.} The Regulatory Compliance Adapter (Pattern~7) wraps all external-facing operations: data access requests, model deployment decisions, output generation, and audit reporting. Adapters are consulted before any operation that crosses the system boundary.

\begin{figure}[htbp]
\centering
\resizebox{\textwidth}{!}{%
\begin{tikzpicture}[
    layer/.style={draw, rounded corners=3pt, minimum width=12cm, minimum height=1.5cm, fill=#1!8, font=\small},
    pattern/.style={draw, rounded corners=2pt, fill=white, font=\footnotesize, minimum height=0.8cm, text width=2.8cm, align=center},
    lbl/.style={font=\scriptsize\bfseries, anchor=north west},
    arr/.style={-{Stealth[length=2.5mm]}, thick, gray!70},
    every node/.style={inner sep=3pt}
]

\node[layer=orange, minimum height=1.2cm] (L5) at (0, 8.0) {};
\node[lbl] at ([xshift=2pt, yshift=-2pt]L5.north west) {Layer 5: Compliance};
\node[pattern, fill=orange!15] (P7) at (1.5, 8.0) {P7: Regulatory\\Compliance Adapter};

\node[layer=purple, minimum height=1.2cm] (L4) at (0, 6.4) {};
\node[lbl] at ([xshift=2pt, yshift=-2pt]L4.north west) {Layer 4: Monitoring};
\node[pattern, fill=purple!15, text width=4cm] (P3) at (1.5, 6.4) {P3: Drift Sentinel\\{\tiny (Type A/B/C classification)}};

\node[layer=red, minimum height=1.2cm] (L3) at (0, 4.8) {};
\node[lbl] at ([xshift=2pt, yshift=-2pt]L3.north west) {Layer 3: Training};
\node[pattern, fill=red!12] (P5) at (1.5, 4.8) {P5: Reification\\Circuit Breaker};

\node[layer=green, minimum height=1.5cm] (L2) at (0, 3.1) {};
\node[lbl] at ([xshift=2pt, yshift=-2pt]L2.north west) {Layer 2: Storage};
\node[pattern, fill=green!12] (P4) at (-0.5, 3.1) {P4: Dual-Ontology\\Layer};
\node[pattern, fill=green!12] (P2) at (3.5, 3.1) {P2: Dormancy-Aware\\Pipeline};

\node[layer=blue, minimum height=1.5cm] (L1) at (0, 1.2) {};
\node[lbl] at ([xshift=2pt, yshift=-2pt]L1.north west) {Layer 1: Ingestion};
\node[pattern, fill=blue!12] (P1) at (-0.5, 1.2) {P1: Ontological\\Checkpoint};
\node[pattern, fill=blue!12] (P6) at (3.5, 1.2) {P6: Terminology\\Version Gate};

\node[font=\footnotesize, below=0.3cm of L1] (ext) {External Data (EHR, KV, SNOMED, ICD)};

\draw[arr] (ext.north) -- (L1.south);
\draw[arr] (P1.north) -- (P4.south);
\draw[arr] (P6.north) -- (P2.south);
\draw[arr] (P4.north) -- ([xshift=-0.8cm]P5.south);
\draw[arr] (P2.north) -- ([xshift=0.8cm]P5.south);
\draw[arr] (P5.north) -- (P3.south);

\draw[arr, dashed, purple!60] (L4.east) -- ++(0.6,0) |- (L1.east)
    node[pos=0.25, right, font=\tiny, text=purple!70, text width=1.2cm, align=left] {monitors\\all layers};

\draw[arr, dashed, orange!60] (L5.west) -- ++(-0.6,0) |- (L1.west)
    node[pos=0.25, left, font=\tiny, text=orange!70, text width=1.2cm, align=right] {wraps all\\external ops};

\draw[arr, red!50, dashed] (P5.east) -- ++(1.8,0) |- (P1.east)
    node[pos=0.25, right, font=\tiny, text=red!60] {feedback path};

\end{tikzpicture}%
}
\caption{Five-layer reference architecture composing all seven ontology-aware design patterns. Solid arrows indicate primary data flow from ingestion through training. Dashed arrows indicate cross-cutting concerns: the Drift Sentinel monitors all layers, the Regulatory Compliance Adapter wraps all external-facing operations, and the feedback path (monitored by the Reification Circuit Breaker) connects model outputs back to data ingestion.}
\label{fig:architecture}
\end{figure}
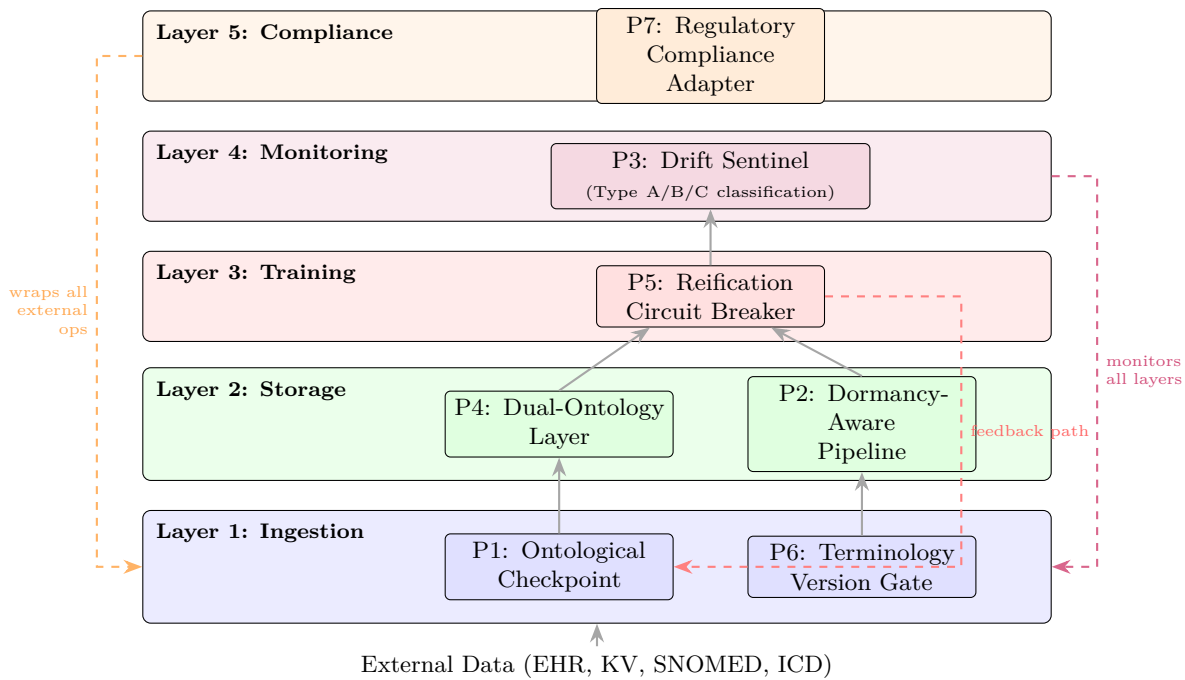

Figure~\ref{fig:architecture} illustrates the five-layer composition and the primary data flow between patterns.

\subsection{Pattern Interaction Matrix}

Table~\ref{tab:interaction} summarises the dependencies and interactions between patterns.

\begin{table}[htbp]
\centering
\caption{Pattern interaction matrix. Arrows indicate direction of dependency or data flow.}
\label{tab:interaction}
\scriptsize
\setlength{\tabcolsep}{2.5pt}
\begin{tabular}{@{}l c c c c c c c@{}}
\toprule
 & \textbf{P1} & \textbf{P2} & \textbf{P3} & \textbf{P4} & \textbf{P5} & \textbf{P6} & \textbf{P7} \\
\midrule
\textbf{P1} & -- & Feeds & Feeds & Feeds & Feeds & Depends & Wrapped \\
\textbf{P2} & Receives & -- & Receives & Stored in & -- & Tagged & Wrapped \\
\textbf{P3} & Monitors & Triggers & -- & Monitors & Triggers & Triggers & Wrapped \\
\textbf{P4} & Receives & Hosts & Monitored & -- & Provides & Gated & Wrapped \\
\textbf{P5} & Observes & -- & Triggered & Uses & -- & -- & Wrapped \\
\textbf{P6} & Required & Tags & Interacts & Gates & -- & -- & Wrapped \\
\textbf{P7} & Wraps & Wraps & Wraps & Wraps & Wraps & Wraps & -- \\
\bottomrule
\end{tabular}
\end{table}

The most critical interaction path is: Ontological Checkpoint (P1) feeds Dual-Ontology Layer (P4), which is monitored by Drift Sentinel (P3), which triggers Reification Circuit Breaker (P5). This chain addresses the full lifecycle from data ingestion distortion through feedback loop prevention.

\subsection{Walkthrough: Primary Care Diabetes Risk Prediction}

To illustrate the patterns in composition, we trace a concrete scenario: a primary care AI system for Type~2 diabetes risk prediction, operating on ICD-10-GM coded encounter data from German ambulatory care.

\textbf{Step 1: Data Ingestion (Patterns 1 and 6).} A batch of 50,000 encounter records arrives from a Kassen\"{a}rztliche Vereinigung data extract. The Terminology Version Gate (P6) checks that all records are coded under ICD-10-GM 2025; 3,200 records from a late-reporting practice are coded under ICD-10-GM 2024 and are routed to the version reconciliation module. The Ontological Checkpoint (P1) computes coding fidelity scores. Records with ICD E11.9 (``Type 2 diabetes mellitus without complications'') from practices with unusually high E11.9 rates receive lower fidelity scores, flagging possible administrative simplification (E11.9 used as a catch-all rather than the more specific E11.65 or E11.69).

\textbf{Step 2: Storage (Patterns 4 and 2).} Records enter the Dual-Ontology Layer (P4). The administrative layer stores codes as received. For records with low fidelity scores, the clinical layer is populated probabilistically: a classification model estimates the most likely specific diabetes subtype from co-occurring codes (HbA1c lab results, medication prescriptions, complication indicators). The Dormancy-Aware Pipeline (P2) identifies 47 records coded with E13 (``Other specified diabetes mellitus''), a rare code in this population. These are routed to the dormant feature store with activation conditions: reactivate if E13 prevalence exceeds 0.5\% in any quarterly cohort or if a domain transfer to a specialised endocrinology context is requested.

\textbf{Step 3: Training (Pattern 5).} The diabetes risk model is scheduled for quarterly retraining. The Reification Circuit Breaker (P5) checks the AI influence ratio: in the previous quarter, the model's risk predictions were displayed in the EHR, and 12\% of new diabetes screening referrals were initiated following a model alert. These referrals generated new encounter data now entering the training set. The AI influence ratio is 12\%, below the 15\% threshold, so retraining proceeds. However, the circuit breaker logs a warning: the ratio has increased from 4\% to 12\% over three quarters, suggesting an upward trend that may breach the threshold by the next cycle.

\textbf{Step 4: Monitoring (Pattern 3).} The Drift Sentinel (P3) detects a distributional shift in E11.65 (``Type 2 diabetes mellitus with hyperglycaemia'') codes over the past two quarters. The semantic fingerprint analysis classifies this as Type~B (administrative) drift: the shift correlates with a regional billing guideline change that encouraged more specific diabetes coding, not with a genuine increase in hyperglycaemic episodes. The model team is alerted to re-evaluate whether the current model's training data distribution still represents the population it will be applied to.

\textbf{Step 5: Compliance (Pattern 7).} The retrained model is submitted for deployment. The Regulatory Compliance Adapter (P7) applies the EU AI Act adapter: the model is classified as high-risk (health domain, clinical decision support). The adapter checks transparency requirements (model card present, training data documentation complete), human oversight provisions (physician-in-the-loop for all risk predictions above the 90th percentile), and EHDS secondary use compliance (data access authorisation valid, purpose limitation documented). The MDR adapter determines that the system does not meet the definition of a medical device under current guidance, but flags that this assessment should be reviewed if the system's outputs are used to directly initiate treatment. Deployment proceeds with conditions.

\textbf{Where conventional pipelines typically fall short.} A clinical AI pipeline without these patterns would likely ingest all records at face value (missing Step~1's fidelity annotation), store only administrative codes (missing Step~2's clinical layer and dormant store), retrain on AI-influenced data without monitoring (missing Step~3's circuit breaker check), miss the billing guideline drift (missing Step~4's semantic classification), and hard-code a single regulatory check that may not cover the AI Act's evolving requirements (missing Step~5's adapter architecture).


\section{Discussion}

\subsection{Trade-offs and Costs}

The seven patterns described in this paper add architectural complexity. This is a deliberate trade-off: complexity at the architecture level in exchange for reduced risk at the clinical output level. However, the trade-off is not universally favourable.

\textbf{Complexity overhead.} A system implementing all seven patterns requires substantially more infrastructure than a conventional ML pipeline. For a well-resourced health-AI company, this overhead may be justified by the risk reduction it provides. For a startup with limited engineering resources, selective adoption is more realistic. We suggest Patterns~1 (Ontological Checkpoint), 3 (Drift Sentinel), and 5 (Reification Circuit Breaker) as a ``minimum viable ontological safety net'' for resource-constrained teams.

\textbf{Latency.} Patterns~1 and 6 add processing time at the data ingestion boundary. For batch-processing pipelines (common in population health analytics), this is unlikely to be consequential. For real-time clinical decision support systems, the added latency may be clinically relevant and requires benchmarking in the specific deployment context.

\textbf{Ontological over-specification.} There is a risk that engineering teams, having adopted an ontological perspective on clinical data, may over-engineer the ontological layer in contexts where simpler statistical approaches would suffice. Not every AI application requires a dual-ontology layer; a mortality prediction model operating on structured lab values may face minimal coding distortion. These patterns address data-layer distortion; they do not address bias in model architecture, training procedures, or deployment context.

\subsection{When Competitors Are Better}

The patterns proposed here complement, rather than replace, existing frameworks. In several dimensions, existing work provides more actionable guidance:

\textbf{Deployment-phase risk management.} \citet{Wiens2019} provides a comprehensive ``do no harm'' framework for ML in healthcare that addresses deployment risks (alert fatigue, automation bias, liability) with specificity that this paper does not attempt. Teams deploying clinical AI should consult Wiens et al.\ for deployment-phase guidance and this paper for data-layer resilience.

\textbf{End-user interface design.} \citet{Sendak2020} provides directly applicable patterns for presenting ML model information to clinical end users, including uncertainty communication and explanation interfaces. This paper does not address the human-computer interaction layer.

\textbf{Statistical drift handling.} ADAPT \citep{Xiong2026} provides a validated statistical approach to handling ICD coding transitions through adversarial transfer learning. For teams primarily concerned with coding transition robustness and who are comfortable treating drift as a statistical rather than ontological phenomenon, ADAPT may be more immediately implementable than our Drift Sentinel pattern.

\textbf{Fairness-first applications.} For applications where algorithmic fairness is the primary concern, \citet{Rajkomar2018} and recent comprehensive reviews \citep{BiasMitigation2025} provide frameworks that address equity dimensions this paper does not.

Our patterns operate at the data ontology layer, upstream of deployment, presentation, drift compensation, and fairness concerns. They are best understood as a foundation upon which these other frameworks can be more effectively applied.

\subsection{Implications for Health-AI Startups}

Startups building clinical AI on electronic health record data should assume that ontological distortion exists until proven otherwise. The evidence from prior work \citep{Stummer2026a, Stummer2026c} suggests that the distortion is structural, not incidental: it arises from the fundamental design of documentation and billing workflows.

Practical recommendations for engineering teams:

\begin{enumerate}
    \item \textbf{Start with Pattern 1 (Ontological Checkpoint).} Even a simple implementation that flags codes from institutions with unusual coding patterns provides value.
    \item \textbf{Add Pattern 3 (Drift Sentinel) before scaling.} As data volumes grow and models are retrained, semantic drift detection becomes essential.
    \item \textbf{Implement Pattern 5 (Reification Circuit Breaker) before closing the loop.} If the AI system's outputs influence clinical documentation in any way, the feedback path must be monitored.
    \item \textbf{Pattern 7 (Regulatory Compliance Adapter) is essential for EU market access.} The AI Act's high-risk classification for clinical AI makes compliance non-optional.
    \item \textbf{Patterns 2, 4, and 6 may be deferred} depending on the specific use case and data sources, but should be on the technical roadmap.
\end{enumerate}

The PRAXIS-AI framework \citep{StummerPRAXIS} can guide which patterns to prioritise based on the specific implementation context and the regulatory landscape of the target market.

\subsection{Connection to Prior Work}

This paper translates theoretical analyses of ontological distortion into an engineering vocabulary. The intended reading path for an engineering audience is: consult \citet{Stummer2026a} for the mechanism of documentary distortion, \citet{Stummer2026c} for the formalisation of AI-driven amplification, and this paper for the architectural response.


\section{Limitations}

We identify seven limitations of this work. These are structural constraints, not incidental shortcomings; they define the boundaries of what this paper claims.

\begin{enumerate}
    \item \textbf{No empirical validation.} The patterns are derived from theoretical analysis of documentary distortion, not from observed failures or measured improvements in production clinical AI systems. Production deployment studies are needed to validate their effectiveness and measure their overhead. \emph{Mitigation:} We identify four patterns with no known precedent and three with partial precedent, providing a clear research agenda for empirical work.

    \item \textbf{No runtime benchmarks.} We do not report latency, storage, compute costs, or throughput impacts for any pattern. The reference architecture is described at a level of abstraction that allows implementation in multiple technology stacks; benchmarks would be stack-specific and deployment-context-dependent. \emph{Mitigation:} The walkthrough in Section~5.3 identifies the processing stages where overhead accumulates, enabling targeted benchmarking.

    \item \textbf{Single-author pattern extraction.} Patterns were extracted by the author of the underlying theoretical work, creating a risk of confirmation bias. The patterns may over-fit the theoretical narrative while missing distortion mechanisms not captured by the prior analyses. Independent extraction by software engineers who have not read the theory would test robustness. \emph{Mitigation:} All patterns are published under CC BY 4.0 and are specified with sufficient detail for independent implementation and critique.

    \item \textbf{EU regulatory scope.} Pattern~7 (Regulatory Compliance Adapter) is described for the EU regulatory context (AI Act, MDR, EHDS). The US (FDA, 21st Century Cures Act), Canadian (Health Canada), UK (MHRA), and LMIC regulatory contexts are not addressed. \emph{Mitigation:} The adapter pattern is explicitly designed for jurisdictional pluggability; the pattern structure is jurisdiction-agnostic even if the described instance is EU-specific.

    \item \textbf{German ambulatory care bias.} The clinical examples, the three-forces model, and the diabetes walkthrough draw primarily on the German ambulatory care context (Kassen\"{a}rztliche Vereinigung billing, ICD-10-GM). Hospital systems, US payer-based systems, single-payer systems, and contexts with different documentation cultures may surface different forces that require different patterns or different calibrations of these patterns. \emph{Mitigation:} The pattern language is deliberately abstract; forces should be reassessed per deployment context.

    \item \textbf{No user study.} We do not report whether CTOs, software architects, or clinical AI engineers find the patterns useful, implementable, or appropriately scoped. The patterns reflect a theoretical analysis of what engineering teams should address, not an empirical analysis of what they currently struggle with. \emph{Mitigation:} We frame the patterns as a ``design vocabulary'' to be refined through community use, not as prescriptive standards.

    \item \textbf{Theoretical dependency.} The patterns are only as sound as the theoretical analyses they translate. If the three-forces model \citep{Stummer2026a}, the reification feedback loop \citep{Stummer2026c}, or the strategic dormancy thesis \citep{Stummer2026d} are empirically falsified, the corresponding patterns lose their grounding. \emph{Mitigation:} Each pattern cites its theoretical source explicitly. Falsification of one source invalidates the corresponding pattern, not the entire set. The patterns are modular by design.
\end{enumerate}


\section{Conclusion}

This paper has proposed seven ontology-aware design patterns that translate theoretical insights about clinical data distortion into a software engineering vocabulary. The patterns address the lifecycle of distortion in clinical AI systems: from data ingestion (Ontological Checkpoint, Terminology Version Gate), through storage and feature management (Dual-Ontology Layer, Dormancy-Aware Pipeline), to training feedback monitoring (Reification Circuit Breaker), continuous semantic oversight (Drift Sentinel), and regulatory compliance (Regulatory Compliance Adapter).

The contribution is the pattern language itself: named, reusable architectural abstractions that make ontological distortion a first-class engineering concern rather than an invisible assumption. The patterns are grounded in theoretical analyses of distortion mechanisms from sociological, decision-theoretic, AI-architectural, and governance perspectives. This paper provides the engineering bridge.

These patterns do not solve ontological distortion. They make it visible and manageable. A clinical AI system implementing these patterns will still operate on distorted data, but the distortion will be annotated, monitored, version-tracked, feedback-controlled, and regulatory-checked rather than silently propagated. This is a meaningful improvement, but it is not a cure.

We do not claim that these seven patterns are exhaustive. They address the distortion mechanisms identified in prior theoretical work. Other mechanisms, not captured by the current analysis, may require additional patterns. We invite the clinical AI engineering community to implement, critique, extend, and, where warranted, discard these patterns in the light of production experience.

The intended audience for this paper is the engineering team that builds clinical AI systems. The cited prior work provides the theoretical depth; this paper provides the architectural response. The next step is empirical: implement the patterns, measure their costs and benefits, and report the results.


\subsection*{Anti-Salami Declaration}

This paper contributes a software engineering perspective (design patterns in the Gang-of-Four pattern language) that is absent from the author's prior publications. The cited prior work addresses the sociology of classification \citep{Stummer2026a}, decision theory and feedback loop formalisation \citep{Stummer2026c}, AI reasoning architecture \citep{Stummer2026d}, and implementation science \citep{StummerPRAXIS}. None provides implementable architectural guidance for engineering teams. This paper does not replicate empirical findings or theoretical claims from those works; it translates them into a different disciplinary form.


\section*{Author Contributions (CRediT)}

\textbf{Florian Odi Stummer:} Conceptualization; Methodology; Formal analysis; Writing (original draft); Writing (review and editing).


\section*{Competing Interests}

The author declares no competing interests.


\section*{Data Availability}

No datasets were generated or analysed during this study. All patterns are specified in full within this manuscript and are available under CC BY 4.0.


\end{document}